%% file: main.tex
\documentclass[letterpaper, 10 pt, conference]{ieeeconf}  

\IEEEoverridecommandlockouts                              

\input{preamble.tex}

\input{commands.tex}
\newcommand{\citet}[1]{\cite{#1}}
\newcommand{\citep}[1]{\cite{#1}}

\title{\LARGE \bf \titlelong}

\author{Ken Nakahara$^{1}$, Aleksei Buvailik$^{1}$,
Prokhor Kotov$^{1}$, and Roberto Calandra$^{1}$
\thanks{$^{1}$LASR Lab, Faculty of Computer Science, TU Dresden, Germany.
{\tt\small \{knakahara, rcalandra\}@lasr.org}.}
\thanks{This work was partly supported by the project GeniusRobot (01IS24083) funded by the Federal Ministry of Research, Technology and Space (BMFTR), by BMFTR in the DAAD project 57616814 (\href{https://secai.org/}{SECAI, School of Embedded Composite AI}), by BMFTR under the Robotics Institute Germany (RIG), and by the German Research Foundation (DFG) under the Cluster of Excellence CARE: Climate-Neutral And Resource-Efficient Construction (EXC 3115, project number 533767731).}
}

\begin{document}

\maketitle
\thispagestyle{empty}
\pagestyle{empty}


\begin{abstract}
	\input{0_abstract.tex}
\end{abstract}


\section{Introduction}
\label{sec:introduction}
	
  \input{1_introduction.tex}


\section{Related Work}
\label{sec:related}

	\input{2_related.tex}


\section{Hardware Setup}
\label{sec:hardware}

	\input{3_hardware.tex}


\section{Temporal Multimodal Learning for Dexterous Grasping}
\label{sec:approach}

	\input{4_approach.tex}


\section{Data Collection}
\label{sec:data}

	\input{5_data.tex}


\section{Experimental Results}
\label{sec:result}

	\input{6_result.tex}


\section{Conclusion}
\label{sec:conclusion}

	\input{7_conclusion.tex}






\section*{Acknowledgment}

  \input{99_acknowledgments.tex}


\bibliographystyle{IEEEtran}
\bibliography{paper}


\clearpage
\appendices
\input{100_appendix.tex}


\end{document}

%% file: preamble.tex
\usepackage[english]{babel}

\usepackage{graphicx}
\usepackage{animate}
\usepackage{epsfig} 				
\usepackage{epstopdf}

\usepackage{listings}
\usepackage{color}
\usepackage{nameref}
\usepackage[hidelinks]{hyperref}
\usepackage{amsmath}	 			
\usepackage{amssymb}  				

\usepackage{dsfont}			
\usepackage{mathtools}

\usepackage{epigraph}
\usepackage{lscape}
\usepackage[]{nomencl}				
\usepackage{algorithm}
\usepackage{algorithmic}
\usepackage{multicol}
\usepackage{multirow}
\usepackage{etoolbox}

\usepackage{caption}
\usepackage{subcaption}
\usepackage{wrapfig}

\usepackage{siunitx}

\usepackage{floatflt}

\usepackage{url}

\usepackage{bm}
\usepackage{comment}
\usepackage{pict2e}

%% file: commands.tex
\newcommand{\email}[1]{\href{mailto:#1}{\nolinkurl{#1}}}

\renewcommand{\sec}[1]{Sec.~\ref{#1}}
\newcommand{\fig}[1]{Fig.~\ref{#1}}

\newcommand{\tab}[1]{Table~\ref{#1}}

\newcommand{\titlelong}[0]{Temporal Visuo-Tactile Learning for Dexterous Grasp Stability}

\newcommand{\website}[0]{\url{https://lasr-lab.github.io/dexterous-grasp-stability/}}
\newcommand{\numgrasps}[0]{10{,}000}
\newcommand{\numobjects}[0]{200}
\newcommand{\digittt}[0]{Digit~360}

\newcommand{\bestrelvizsuccessgrasp}[0]{10.5}
\newcommand{\bestrelchancesuccessgrasp}[0]{32.5}
\newcommand{\bestrelrandomsuccessgrasp}[0]{34.5}

%% file: 0_abstract.tex
Humans can grasp everyday objects with almost perfect success rates using fingertip tactile feedback, yet much of the robotic grasping literature emphasizes vision-based grasp selection with parallel grippers.
In this work, we systematically investigate how high-resolution, dynamic tactile sensing contributes to grasp stability prediction and model-guided grasping in dexterous robotic hands.
To this end, we collected a dataset of \numgrasps{} grasp trials across \numobjects{} objects using a multi-fingered robotic hand equipped with four \digittt{} tactile sensors, recording external vision, proprioception, and tactile streams throughout each grasp.
With this dataset, we trained end-to-end temporal multimodal models to predict post-lift stability from pre-lift grasp observations and compared sensing modalities and encoding backbones.
Experimental results and controlled input ablations show that incorporating touch, and particularly high-resolution, dynamic touch, improves grasp stability prediction. 
Finally, we deployed the learned predictor as an online stability gate on the real robot, where visuo-tactile model-guided regrasping improved the success rate among executed lifts by \bestrelvizsuccessgrasp{} percentage points over a non-tactile gate.
These results show how rich fingertip sensing and expressive temporal models that capture the dynamics of touch can support learned grasping with multi-fingered hands without explicit contact or force modeling, providing a scalable data-driven path from tactile experience toward stable dexterous manipulation.
The dataset is publicly available at \website{}.

%% file: 1_introduction.tex
Stable grasping with multi-fingered hands cannot be reduced to selecting a grasp pose from visual observations alone.
It is an inherently interactive problem, where stability depends on how contacts form and contact forces evolve as the hand closes around an object, shares load across multiple fingertips, and begins to lift.
Humans handle these changing contact conditions by combining vision with tactile feedback about fingertip contacts that are difficult to observe externally~\citep{johansson2009coding,li2020review}.
However, much research on robotic grasping has focused on vision-based grasp selection, often for simple parallel grippers~\citep{du2021vision,kleeberger2020survey}.
Recent visuo-tactile learning approaches have shown that tactile input improves grasp outcome prediction compared to vision-only baselines~\citep{calandra2017feeling,calandra2018more,han2024learning}.  
At the same time, modern vision-based tactile sensors are becoming more capable and accessible, enabling richer contact measurements during manipulation~\citep{yuan2017gelsight,ward2018tactip,lambeta2020digit}.  
For multi-fingered grasping, however, it remains unclear which sensing modalities are most informative for stability prediction, how tactile spatial detail and temporal dynamics contribute, and how to model these signals effectively.

\begin{figure}[t]
    \centering
    \includegraphics[width=\linewidth]{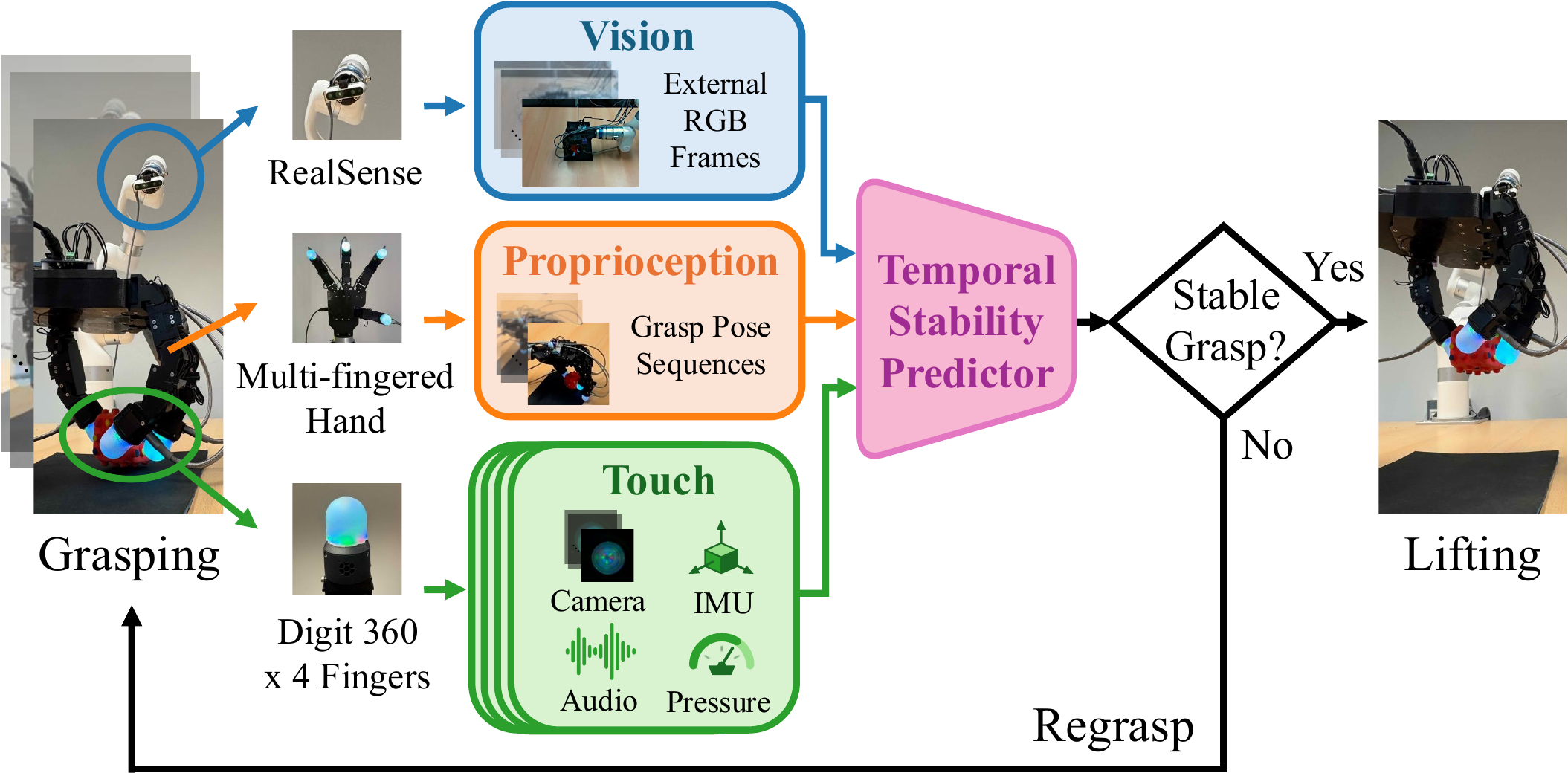}
    \caption{\textbf{Temporal multimodal stability prediction for dexterous grasping.} During grasp execution, the robot observes the object and fingertip contacts through external vision, proprioception, and four \digittt{} tactile sensors.
    A temporal stability predictor fuses these signals to estimate post-lift grasp stability and gate lift-or-regrasp decisions online, leading to higher success among executed lifts.}
    \label{fig:overview}
\end{figure}

In this paper, we study how temporal multimodal sensing supports post-lift stability prediction from pre-lift grasp observations and model-guided grasping with multi-fingered hands, as illustrated in \fig{fig:overview}.
We center our study on grasp stability prediction and practical real-robot deployment: what to sense, which spatial and temporal cues matter, and how to model them to enable stable grasping across diverse objects.
Our contributions are threefold:
\begin{itemize}
    \item We collect a large-scale multimodal dataset of \numgrasps{} dexterous grasp trials across \numobjects{} objects, with continuous recordings of external vision, high-resolution tactile signals from four \digittt{} sensors, and proprioception throughout the full grasping process.
    \item We systematically evaluate dexterous grasp stability prediction across sensing modality combinations and static and temporal encoders under object-disjoint cross-validation. We further use controlled input ablations to isolate the roles of tactile spatial detail and temporal cues in prediction.
    \item We deploy the learned visuo-tactile stability predictor as an online lift-or-regrasp gate on a multi-fingered robotic platform and achieve higher success among executed lifts on unseen objects than a non-tactile gate.
\end{itemize}
These findings provide practical guidance for data-driven dexterous grasping systems and highlight the value of high-resolution, dynamic tactile sensing for robust manipulation.

%% file: 2_related.tex
\subsection{Grasp Stability Prediction}
Grasp stability prediction aims to determine whether a grasp will remain secure after lifting.
Analytical grasp quality measures often assume known contact geometry and force models~\citep{roa2015grasp}, which are difficult to obtain reliably on real hardware.
Tactile grasp controllers provide useful principles, but often rely on hand-tuned tactile-event thresholds and grip adjustment rules~\citep{yamaguchi2017implementing,romano2011human}, motivating learned stability assessment from tactile and robot-state observations~\citep{li2020review, kwiatkowski2017grasp}.
End-to-end models using vision-based tactile sensors have demonstrated the benefits of touch over vision alone for stability prediction and enabled model-guided regrasping~\citep{calandra2017feeling,calandra2018more}.
Subsequent work has explored attention-based visuo-tactile fusion~\citep{cui2020self}, Transformer-based grasping of deformable objects~\citep{han2024learning}, and the addition of sound for stability and gentleness prediction~\citep{nakahara2025learning}.
Tactile stability prediction for multi-fingered hands remains comparatively underexplored, with prior work including Bayesian haptic exploration with fingertip force sensing~\citep{siddiqui2021grasp}.
Our work systematically examines how sensing modalities and tactile spatial and temporal cues contribute to multi-fingered stability prediction, and deploys the learned predictor as an online stability gate.

\subsection{Learning and Control for Tactile Dexterous Grasping}
Multi-fingered hands support diverse contact configurations and enveloping grasps, providing structural advantages for robust grasping.
Despite recent progress in learning-based dexterous grasping, robust real-world execution remains challenging due to high-dimensional hand control and uncertain hand-object contacts~\citep{song2025overview}.
Existing approaches include physics-aware grasp synthesis~\citep{zhong2025dexgrasp} and hand-centric object representations for robust policy learning~\citep{zhang2025robustdexgrasp}.
Tactile sensing complements these approaches by providing local contact feedback during multi-fingered grasping.
Such feedback has enabled adaptive visual-tactile grasp refinement~\citep{yan2024avita}, paper picking using diffusion policies~\citep{lin2025pptac}, gentle grasping with anthropomorphic hands~\citep{ford2023tactile}, and grasp adaptation using full-hand tactile sensing~\citep{zhao2025embedding}.
Building on this line of dexterous tactile grasping research, we use rich tactile sensing alongside vision and proprioception to predict post-lift stability from temporal pre-lift observations.

\subsection{Vision-Based Tactile Sensors}
Vision-based tactile sensors typically capture deformation of a compliant contact surface as images~\citep{shimonomura2019tactile}, enabling the use of standard image-processing methods and deep visual encoders.
Representative examples include GelSight~\citep{yuan2017gelsight} for high-resolution contact geometry and TacTip~\citep{ward2018tactip} for biomimetic soft optical fingertip sensing.
Compact and low-cost sensors such as DIGIT~\citep{lambeta2020digit} have been used to learn visuo-tactile prehensile manipulation with multi-fingered hands (e.g., \cite{nakahara2025learning,suresh2024neuralfeels}).
\digittt{}~\citep{lambeta2024digitizing} extends this line to a multimodal fingertip, combining an internal fisheye camera for high-resolution tactile imaging across a hemispherical contact surface with additional sensing modalities such as audio and pressure.
Recent work has used \digittt{} to learn multisensory touch representations from roughly one million contact-rich interactions~\citep{higuera2025tactile} and visuo-tactile world models for contact-aware prediction and planning~\citep{higuera2026visuotactile}.
We use four \digittt{} fingertips to investigate how rich, dynamic tactile sensing supports learned dexterous grasping.

%% file: 3_hardware.tex
\begin{figure}[t]
    \centering
    \includegraphics[width=\linewidth]{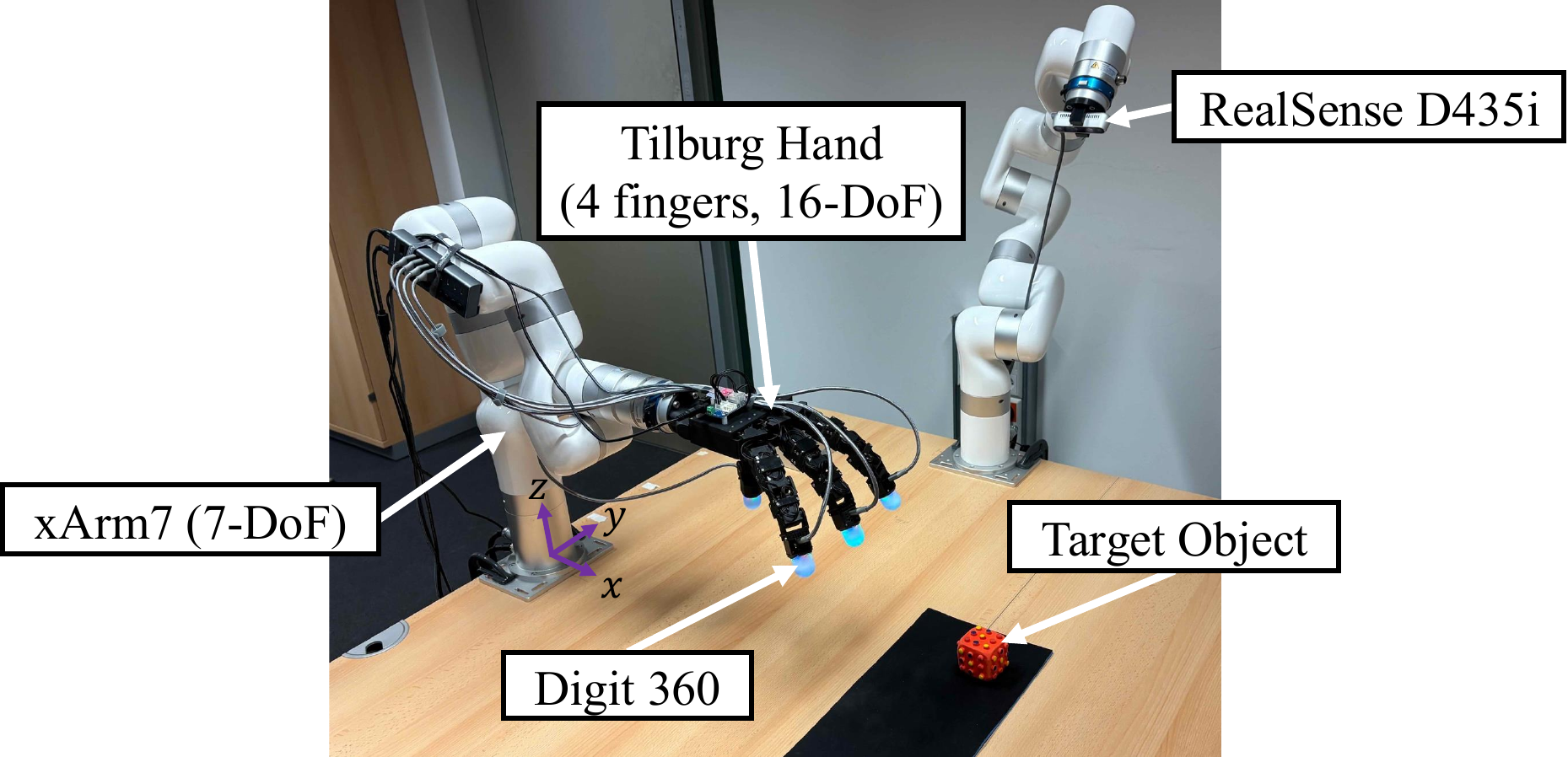}
    \caption{Hardware setup. The system continuously records external visual observations, proprioception, and tactile streams from four \digittt{} sensors throughout each grasp trial.}
    \label{fig:hardware_setup}
\end{figure}

Our experimental platform is shown in \fig{fig:hardware_setup}.
The grasping arm is a 7-DoF xArm7 equipped with a right-handed Tilburg Hand~\citep{tilburgRoboticsHand}, which has four fingers with four actuated joints each (16 DoF in total).
We mount a \digittt{} sensor~\citep{lambeta2024digitizing} on each fingertip and record its tactile images, audio, inertial measurements, and pressure readings.
A RealSense D435i RGB-D camera, mounted on another stationary arm, provides an external view of the workspace.
With this setup, the robot grasps an object placed randomly within a \SI{200}{\milli\meter}$\times$\SI{400}{\milli\meter} tabletop workspace, with external visual observations, proprioceptive signals from the hand and arm, and fingertip tactile streams recorded continuously throughout each grasp trial.

%% file: 4_approach.tex
\begin{figure*}[t]
    \centering
    \includegraphics[width=\textwidth]{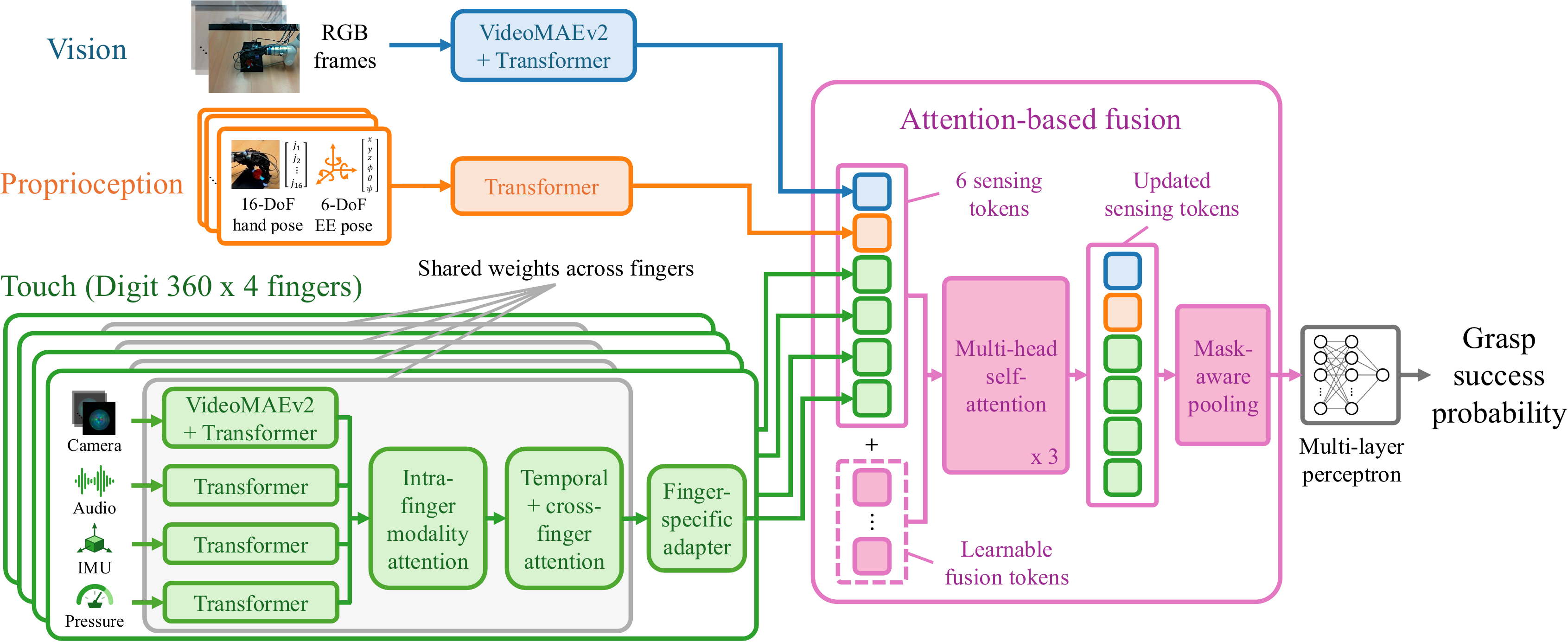}
    \caption{Model architecture for temporal multimodal grasp stability prediction. An observation window from the pre-lift grasp phase, containing RGB frames, proprioception, and tactile streams from four \digittt{} sensors, is encoded by modality-specific Transformer-based branches. Finger-level tactile features are integrated across modalities, time, and fingers, then combined with RGB and proprioceptive features through attention-based fusion to predict post-lift grasp stability.}
    \label{fig:model_architecture}
\end{figure*}

We formulate grasp stability prediction as an end-to-end learning problem~\citep{calandra2017feeling}, using observations from the pre-lift grasp phase to predict post-lift stability.
We deploy the learned predictor online as a stability gate: the robot lifts only when the predicted stability is sufficiently high; otherwise, it releases its grip and attempts a regrasp.

\subsection{End-to-End Grasp Stability Prediction}
\label{sec:end2end_prediction}

Given a sensory observation window $\mathbf{s}_{t:t+T}$ from the pre-lift grasp phase, we learn a function that predicts the probability of a successful grasp $f_{\theta}(\mathbf{s}_{t:t+T})=p(o=1 \mid \mathbf{s}_{t:t+T})\in[0,1]$, where the stability label $o=1$ indicates that the object is successfully lifted and held without being dropped.
We parametrize $f_{\theta}$ with a temporal multimodal neural network that encodes external RGB frames, proprioception, and multimodal fingertip tactile streams from this observation window.
We train the network to classify stable and unstable grasps by minimizing binary cross-entropy loss on the dataset of \numgrasps{} grasp trials described in \sec{sec:data}.

\textbf{Network Design.}
The architecture shown in \fig{fig:model_architecture} takes a three-second pre-lift observation window anchored at grasp initiation and separately encodes vision, proprioception, and touch before fusing the resulting features for stability prediction.
We encode 16-frame external RGB and tactile-camera clips using pretrained VideoMAEv2-Base backbones~\citep{wang2023videomaev2} followed by temporal Transformers, with the tactile encoder shared across fingers.
All visuo-tactile frames are resized to $256 \times 256$ and cropped to $224 \times 224$ pixels, with consistent crop coordinates across frames within each clip.
Training uses one center-cropped and one randomly cropped version of each sample, doubling the number of training samples per epoch; evaluation uses center crops.
Transformer-based sequence encoders process the remaining temporal streams: proprioception is represented as a 100-step sequence comprising 16 hand joint angles and a 6-D end-effector pose, while per-fingertip audio, IMU, and pressure are represented as 224-step sequences of 256-bin log-magnitude spectra, three-axis acceleration, and scalar pressure readings, respectively.
To account for unavailable sensing streams, we use modality masks during touch fusion and cross-modal aggregation.
For the touch branch, encoded camera, audio, IMU, and pressure features are resampled to a common temporal grid and combined at each time step by a modality-fusion Transformer with weights shared across fingers.
The resulting four-finger temporal tokens are processed by weight-shared factorized Transformer blocks over time and fingers, followed by lightweight finger-specific residual adapters to capture finger-dependent biases, and then pooled over time into one token per fingertip.
To combine the sensing streams, the RGB, proprioceptive, and four fingertip tokens are concatenated with 16 learnable fusion tokens in a shared 512-dimensional space and jointly processed by three Transformer layers with eight attention heads.
After mask-aware pooling over the updated branch tokens, the fused representation is fed to a two-layer MLP prediction head with a 512-dimensional hidden layer, layer normalization, GELU activation, and a dropout rate of 0.2, followed by a sigmoid to produce the grasp success probability.

\subsection{Model-Guided Grasp and Regrasp Execution}
\label{sec:regrasp_optimization}

As depicted in \fig{fig:overview}, we deploy the trained stability predictor as an online gate for deciding whether to lift or regrasp.
Following the initialization procedure in \sec{sec:data}, the robot reaches the vicinity of the object and closes the hand toward a sampled grasp configuration.
The predictor evaluates a three-second observation window beginning at grasp initiation, comprising external RGB frames, proprioception, and tactile streams from the four \digittt{} sensors.
The robot lifts only when the predicted grasp success probability exceeds a predetermined threshold; otherwise, it opens the hand, applies a small relative displacement to the end-effector, and attempts another sampled grasp.

%% file: 5_data.tex
We collect a dataset of \numgrasps{} multi-fingered grasp trials under randomized grasp conditions, using \numobjects{} objects of varying sizes, shapes, and materials (\fig{fig:object_set}).
As shown in \fig{fig:data_collection_process}, our data collection procedure consists of object localization, reaching, grasping, lifting, automatic stability labeling, and resetting.
At the start of each trial, the target object is placed at a random position in a \SI{200}{\milli\meter}$\times$\SI{400}{\milli\meter} workspace, with $x\in\SI[parse-numbers=false]{[650,850]}{\milli\meter}$ and $y\in\SI[parse-numbers=false]{[-200,200]}{\milli\meter}$ in the grasping-arm base frame, as depicted in \fig{fig:hardware_setup}.
We use SAM~3~\citep{carion2026sam3} to segment the object in the external RGB image from a text prompt, then estimate its center in the arm base frame from the mask centroid, depth image, and calibrated camera extrinsics.
Given this detected object position, we randomize the target position of the end-effector, defined at the base of the hand's middle finger, by adding translational perturbations, sampled by default from \SI[parse-numbers=false]{[-50,50]}{\milli\meter} in $x$ and $y$ and \SI[parse-numbers=false]{[-25,25]}{\milli\meter} in $z$ but adjusted when needed for object size and dataset balance.
The end-effector yaw is also sampled from $[-20^\circ,20^\circ]$.
The robot moves the end-effector to the sampled target and samples the 16 hand joint targets independently and uniformly from calibrated joint-wise ranges.
These ranges span $15$--$20^\circ$ for most joints and $45^\circ$ for the index, middle, and ring MCP joints.
The hand then closes using 50 position-control waypoints linearly interpolated between its measured and target configurations, with commands issued at a nominal \SI{50}{\hertz}.
After hand closure and completion of the observation window, the robot directly lifts the object by \SI{100}{\milli\meter}, unlike during model-guided deployment (\sec{sec:regrasp_optimization}). 
If the object is lifted and held for \SI{3}{\second} without dropping, the trial is labeled $o=1$ (success); otherwise, $o=0$ (failure).
We automate stability labeling based on the object's post-lift height detected by SAM~3, with proprioceptive fingertip-displacement cues used only when detection fails, e.g., under occlusion.
While most labels were generated through this self-supervision, some were manually corrected based on visual inspection.
The robot then automatically returns to its initial position for the next trial, first randomly relocating the object within the workspace after a successful grasp.

Throughout each trial, we continuously and asynchronously record external RGB frames ($640\times480$), proprioception (16 hand joint angles and a 6-D end-effector pose), and tactile images ($690\times690$) at \SI{30}{\hertz}, along with \SI{48}{\kilo\hertz} stereo audio, three-axis acceleration, and pressure from each \digittt{} sensor, as well as commanded arm and hand actions and phase annotations.
For stability prediction, we extract a \SI{3}{\second} window starting at grasp initiation and ending before lift onset, and resample the streams into synchronized model inputs described in \sec{sec:end2end_prediction}.
The dataset is nearly balanced, with 47.1\% stable and 52.9\% unstable trials.

%% file: 6_result.tex
We evaluate the proposed temporal multimodal stability predictor in both offline and online settings.
First, we analyze how sensing modalities, backbone design, dataset size, tactile spatial resolution, temporal observations, and finger-specific inputs affect offline prediction accuracy.
Second, we deploy the learned predictor as an online stability gate on unseen objects and compare it with a non-tactile gate, a direct-lift baseline, and a random gate.
For all experiments, we train each model for 100 epochs with AdamW, using a batch size of 4, an initial learning rate of $10^{-6}$, weight decay of $10^{-4}$, and cosine scheduling with a 0.03 warmup ratio.

\begin{figure}[t]
    \centering
    \includegraphics[width=\linewidth]{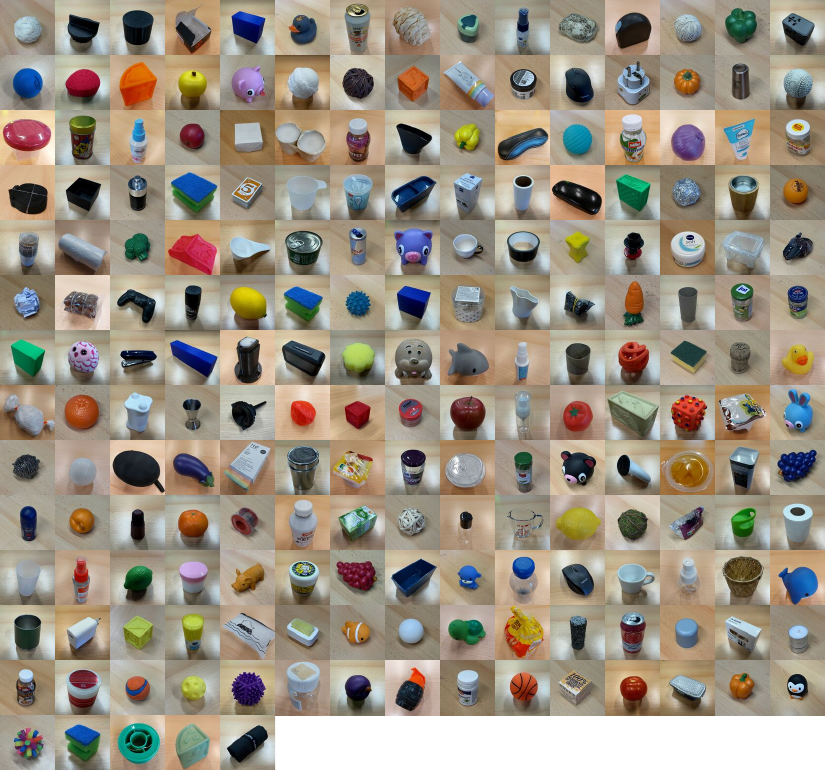}
    \caption{The \numobjects{} objects used to collect \numgrasps{} grasping trials.}
    \label{fig:object_set}
\end{figure}

\begin{figure}[t]
    \centering
    \includegraphics[width=\linewidth]{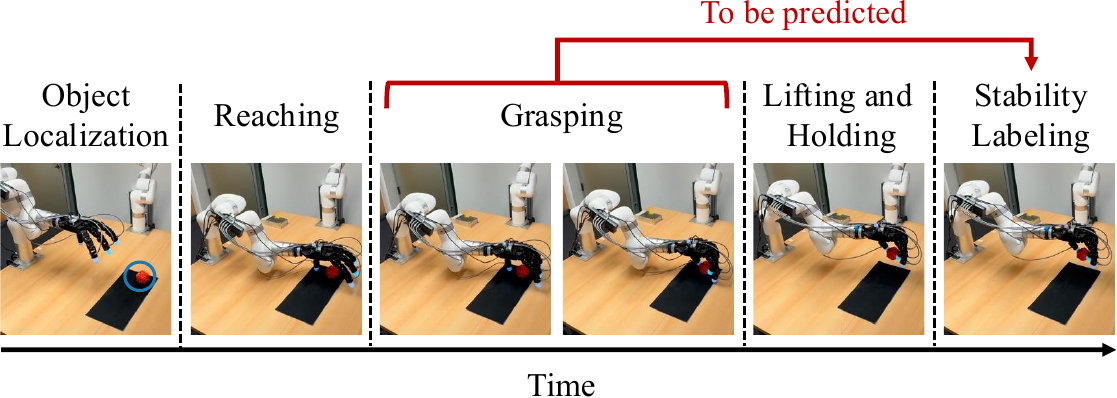}
    \caption{Automated self-supervised data collection timeline. Each trial comprises object localization, randomized reaching and grasping, lifting, and stability labeling from object height changes, with continuous multimodal recording.}
    \label{fig:data_collection_process}
\end{figure}

\subsection{Offline Evaluation of Grasp Stability Prediction}
\label{sec:cross_validation}

We compare models using 5-fold object-disjoint cross-validation (CV) on the dataset of \numgrasps{} grasp trials described in \sec{sec:data}.
Each fold holds out all trials from a subset of objects, thereby measuring generalization to unseen object instances.
We compute classification accuracy from the final-epoch checkpoint by thresholding the predicted grasp success probability at $0.5$ and report its mean and standard deviation (SD) across the five folds.
For each input ablation, we train a separate model with the corresponding input modification applied during both training and evaluation.

\begin{figure*}[t]
    \centering
    \includegraphics[width=\textwidth]{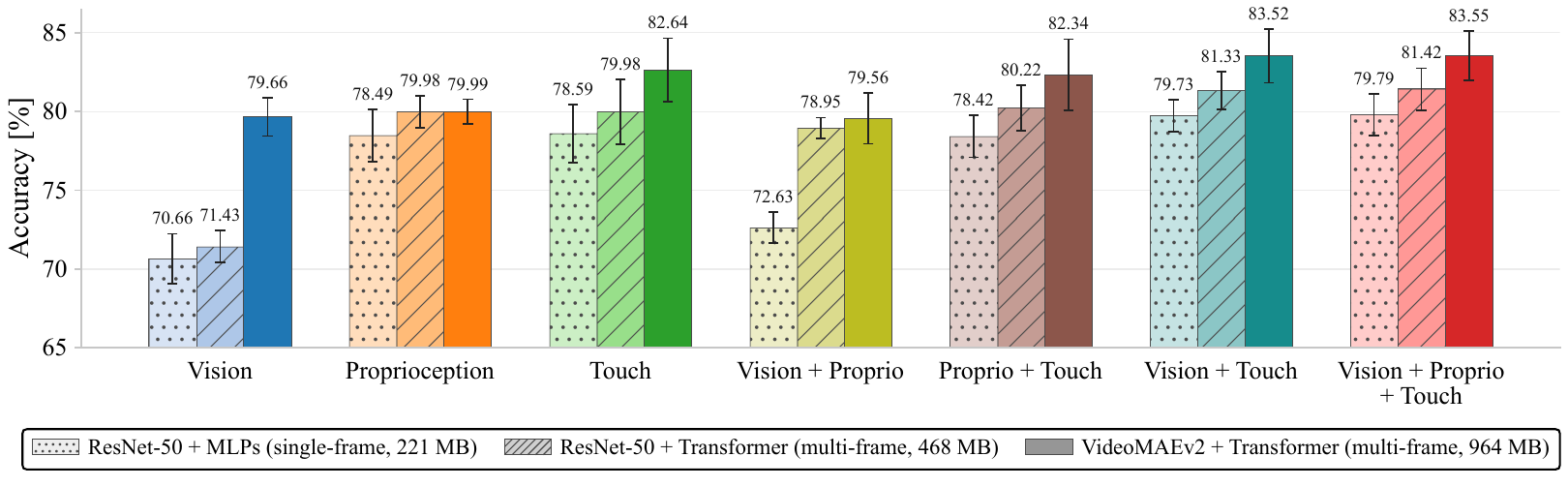}
    \caption{Mean accuracy with SD error bars for modality and backbone ablations under 5-fold CV. Touch-enabled models achieve the strongest performance. Multi-frame models consistently outperform the static single-frame baseline, with the VideoMAEv2-based model performing best, suggesting the value of expressive temporal models for capturing grasp dynamics.}
    \label{fig:validation_results}
\end{figure*}

\begin{figure*}[t]
    \centering
    \includegraphics[width=\textwidth]{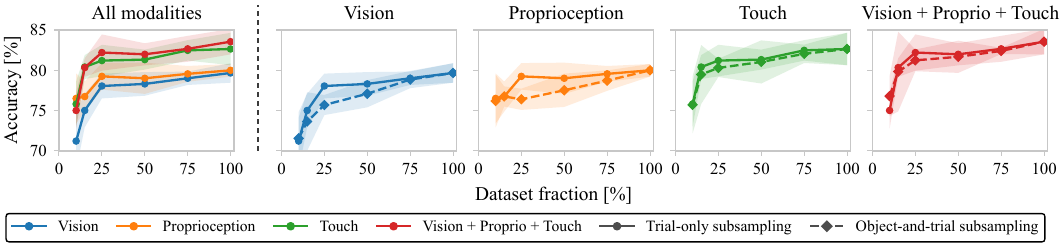}
    \caption{Mean accuracy with SD shading across data scales for VideoMAEv2-based models. Trial-only subsampling (solid) reduces trials per object while retaining the object set; object-and-trial subsampling (dashed) also reduces the number of objects by the same ratio. The leftmost panel compares modalities; the others compare subsampling protocols. Accuracy generally improves with more data, while touch-enabled models remain comparatively robust to reduced object coverage.}
    \label{fig:data_amount_analysis}
\end{figure*}

\subsubsection{Effect of sensing modalities}
As shown in \fig{fig:validation_results}, touch-enabled models lead the VideoMAEv2-based modality ablation: touch-only, vision--touch, and vision--proprioception--touch achieve mean accuracies of $82.64\%$, $83.52\%$, and $83.55\%$, respectively.
Removing touch from the vision--proprioception--touch model reduces accuracy by $3.99$ percentage points to $79.56\%$, supporting the value of local contact cues in this setup.
The vision--proprioception--touch model exceeds touch-only by a modest margin of $0.91$ percentage points (paired $t$-test across five matched folds, $p=0.035$), while its accuracy nearly matches that of vision--touch, suggesting limited additional benefit from proprioception.
With vision and proprioception retained, using only tactile-camera input in the touch branch achieves $83.41\%$ accuracy, close to $83.55\%$ with all tactile streams.
Using audio, IMU, or pressure as the sole tactile modality yields $79.60$--$79.94\%$, close to the non-tactile baseline ($79.56\%$).

\subsubsection{Effect of encoding backbone design}
We next compare three designs for encoding the grasp observation window.
The VideoMAEv2-based model uses pretrained video backbones for visuo-tactile clips, whereas ResNet-Transformer aggregates pretrained ResNet-50~\citep{he2016deep} frame features with temporal Transformers.
Adapted from the late-fusion design introduced in~\citep{calandra2017feeling} to our preprocessing and four-finger sensing setup, the static ResNet-50+MLPs baseline encodes only the final observation, using ResNet-50 for visuo-tactile images and MLPs for non-image inputs.
For the vision--proprioception--touch models, the static baseline achieves $79.79\%$ accuracy (\fig{fig:validation_results}), indicating that the final pre-lift state is already informative.
The temporal ResNet-Transformer and VideoMAEv2-based models achieve higher mean accuracies of $81.42\%$ and $83.55\%$, respectively, suggesting the value of expressive temporal modeling.

For the VideoMAEv2-based model, replacing raw tactile RGB images with absolute differences from each clip's first frame yields $82.58\%$ accuracy.
Separately, pooling the four fingertip representations into a single touch token before cross-modal fusion yields $83.24\%$.

\subsubsection{Effect of dataset size}
To assess the effect of dataset size, we train and evaluate VideoMAEv2-based predictors on 10\%, 15\%, 25\%, 50\%, 75\%, and 100\% subsets of the \numgrasps{}-trial dataset.
At each fraction, trial-only subsampling retains all objects, whereas object-and-trial subsampling retains the same number of trials from proportionally fewer objects.
As shown in \fig{fig:data_amount_analysis}, accuracy generally increases with dataset size.
Among the single-modality models, vision-only degrades most as data is reduced, while proprioception-only is less sensitive but generally performs below touch-enabled models.
Touch-only and vision--proprioception--touch show comparatively small differences between the two protocols.
The larger accuracy gaps between the two protocols for vision and proprioception at 25\%--50\% suggest a benefit from covering more objects with the same number of trials.

\subsubsection{Effect of tactile spatial resolution}
\begin{figure}[t]
    \centering
    \includegraphics[width=\linewidth]{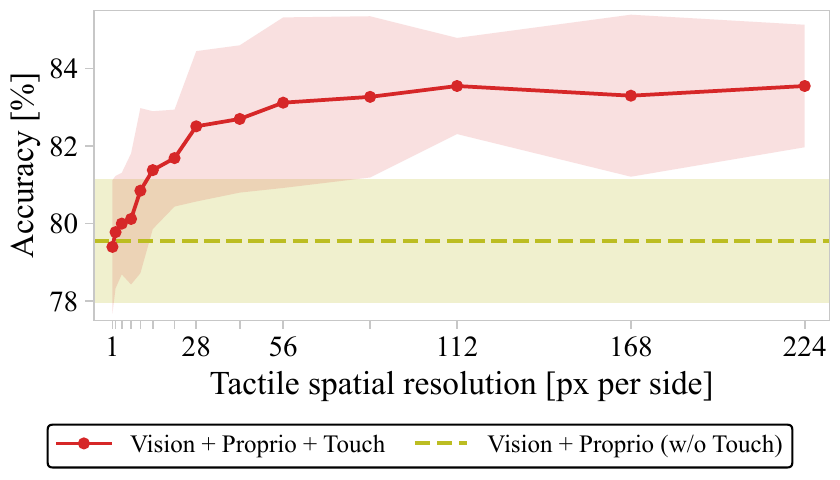}
    \caption{Mean and SD of accuracy across tactile-image spatial resolutions, with the non-tactile baseline shown as a horizontal band. Higher tactile resolution improves accuracy.}
    \label{fig:tactile_spatial_ablation}
\end{figure}

We next examine how the tactile benefit depends on spatial contact detail.
For each reduced resolution ($1$, $2$, $4$, $7$, $10$, $14$, $21$, $28$, $42$, $56$, $84$, $112$, and $168$ pixels per side), we area-downsample the tactile frames and bilinearly restore them to $224\times224$ before encoding, keeping all other inputs and training settings fixed.
As shown in \fig{fig:tactile_spatial_ablation}, the $1\times1$ condition, which retains only each frame's mean RGB values, achieves $79.40\%$ accuracy, close to the non-tactile baseline ($79.56\%$).
Mean accuracy rises steeply up to $28\times28$ ($82.51\%$), then improves more gradually and reaches $83.55\%$ at both $112\times112$ and $224\times224$, supporting the value of spatially resolved contact patterns with diminishing gains at higher resolutions.

\subsubsection{Effect of temporal observations}
\begin{figure}[t]
    \centering
    \begin{subfigure}[b]{0.48\textwidth}
        \centering
        \includegraphics[width=\textwidth]{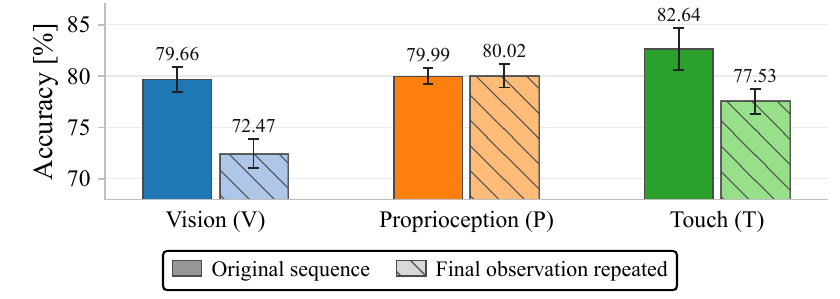}
        \caption{Single-modality models}
        \label{fig:temporal_ablation_a}
    \end{subfigure}
    \begin{subfigure}[b]{0.48\textwidth}
        \centering
        \includegraphics[width=\textwidth]{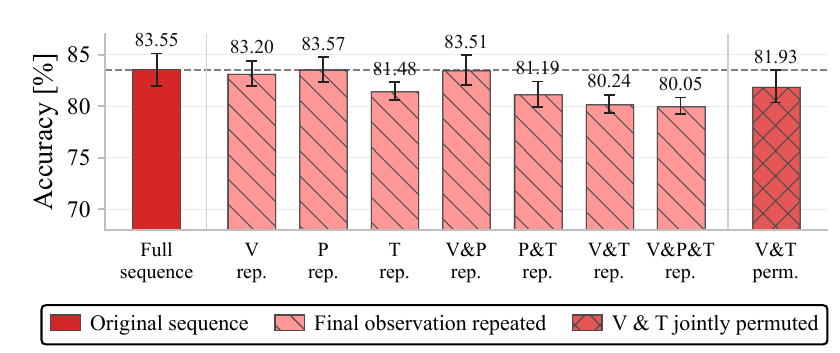}
        \caption{Vision + Proprio + Touch models}
        \label{fig:temporal_ablation_b}
    \end{subfigure}
    \caption{Mean and SD of accuracy under temporal input ablations for (a) single-modality and (b) vision--proprioception--touch models. Final-observation repetition removes temporal variation, while joint vision--touch permutation in (b) disrupts temporal order. Removing tactile dynamics lowers accuracy, highlighting the importance of temporal tactile information.}
    \label{fig:temporal_ablation}
\end{figure}

Since the backbone comparison in \fig{fig:validation_results} also changes architecture, capacity, and pretraining, we examine temporal information through input perturbations while keeping the VideoMAEv2-based architecture and training protocol fixed.
We first repeat each selected stream's final observation throughout the window, preserving the final grasp state while removing within-window variation.
Repetition lowers vision-only and touch-only accuracy but leaves proprioception-only accuracy nearly unchanged (\fig{fig:temporal_ablation_a}).
In the vision--proprioception--touch model (\fig{fig:temporal_ablation_b}), repeating vision or proprioception has little effect, whereas repeating touch lowers accuracy by $2.07$ percentage points; additionally repeating vision or proprioception causes further reductions.
We also apply a common random permutation to the external RGB and four tactile-camera clips in each trial, preserving all frames and alignment among these image streams.
The joint vision--touch permutation in \fig{fig:temporal_ablation_b} lowers accuracy by $1.62$ percentage points, supporting the value of temporal structure beyond frame content.
Overall, these results support the value of temporal tactile information, with multimodal prediction less affected by removing visual than tactile variation.

\begin{figure}[t]
    \centering
    \includegraphics[width=\linewidth]{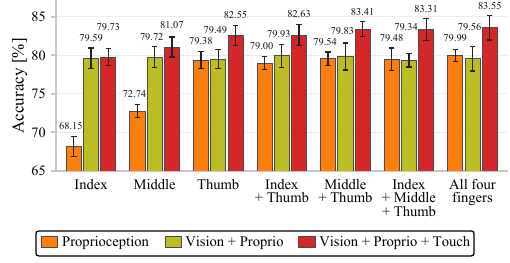}
    \caption{Mean and SD of accuracy under finger-specific input ablations; joint angles and, when enabled, tactile streams are restricted to selected fingers. Vision largely compensates for restricted joint-state inputs, whereas touch, particularly from the thumb as the principal opposing finger, provides stability cues not fully captured by external vision.}
    \label{fig:finger_ablation}
\end{figure}

\subsubsection{Effect of finger-specific inputs}
We restrict the VideoMAEv2-based model's finger-specific joint angles and, when enabled, tactile streams to selected fingers; all grasps are still executed with the full four-fingered hand.
Across finger subsets, mean accuracies span $11.84$ percentage points with proprioception alone, $0.59$ with vision--proprioception, and $3.82$ with vision--proprioception--touch (\fig{fig:finger_ablation}), suggesting that vision compensates for restricted joint-state inputs, while finger-specific tactile cues provide information not fully captured by vision.
Relative to vision--proprioception with the same finger subset, touch improves accuracy by $3.06$ percentage points for the thumb, the principal opposing finger, versus $1.35$ for the middle finger and $0.14$ for the index finger.
The all-finger configuration yields the highest mean accuracy ($83.55\%$; $3.99$-point gain), closely followed by the thumb--middle subset ($83.41\%$).

\begin{figure*}[t]
    \centering
    \includegraphics[width=\textwidth]{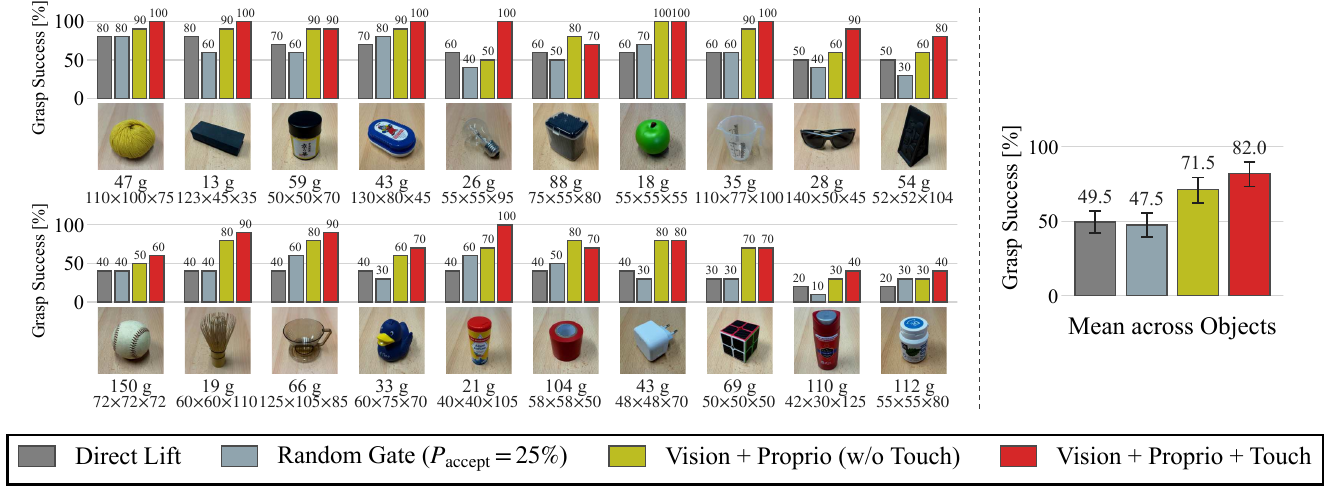}
    \caption{Grasp success rates among executed lifts for four policies on 20 unseen objects. Object mass and dimensions ($W\times D\times H$ in \si{\milli\metre}) are shown below each image. The vision--proprioception--touch gate outperforms the vision--proprioception gate, direct lifting, and the random gate by \bestrelvizsuccessgrasp{}, \bestrelchancesuccessgrasp{}, and \bestrelrandomsuccessgrasp{} percentage points, respectively. Error bars show object-cluster bootstrap $95\%$ confidence intervals for overall success rates; per-object rates are based on 10 executed lifts per policy.}
    \label{fig:online_grasping}
\end{figure*}

\subsection{Real-Robot Grasping Evaluation}

We now evaluate four real-robot grasping policies: tactile (vision, proprioception, and touch) and non-tactile (vision and proprioception) stability gates using VideoMAEv2-based predictors trained on the full \numgrasps{}-trial dataset; a direct-lift baseline; and a random gate.
Each policy is evaluated on 20 unseen objects excluded from the 200-object dataset used for training (\fig{fig:object_set}), with 10 executed lifts per object (\fig{fig:online_grasping}).
Each episode starts with the object randomly placed within the \SI{200}{\milli\meter}$\times$\SI{400}{\milli\meter} workspace, with its resting configuration matching \fig{fig:online_grasping} for consistent comparisons.
All policies use the object-reaching procedure in \sec{sec:data}; initial reaching and regrasp motions use translational perturbations of \SI[parse-numbers=false]{[-25,25]}{\milli\meter} in $x$ and $y$ and \SI[parse-numbers=false]{[-15,15]}{\milli\meter} in $z$, with absolute yaw and local regrasp yaw offsets both limited to $[-20^\circ,20^\circ]$.
The direct-lift baseline always lifts after the initial grasp; both learned gates apply the decision rule in \sec{sec:regrasp_optimization} with an empirically set threshold of $0.95$, fixed before testing.
Below this threshold, the robot releases its grip and samples a regrasp.
Each episode allows one initial grasp and up to five regrasps; six rejections trigger a no-lift reset, returning the robot home and restoring the object's resting configuration if needed.
To assess informed grasp selection, the random gate accepts each candidate with probability $0.25$, approximating the tactile gate's acceptance rate, and follows the same retry and reset protocol.
The tactile, non-tactile, random, and direct-lift policies used 779, 690, 764, and 200 grasp candidates, with 77, 61, 49, and 0 no-lift resets, respectively.
Since no-lift resets for the learned gates are intended abstentions rather than failed lifts, we assess lift-decision reliability primarily by success among executed lifts, while also reporting a conservative episode-level success rate that counts abstentions as failures.

As shown in \fig{fig:online_grasping}, the tactile gate achieves the highest success rate among executed lifts at $82.0\%$ ($164/200$), exceeding the non-tactile and random gates by \bestrelvizsuccessgrasp{} and \bestrelrandomsuccessgrasp{} percentage points, respectively (two-sided object-level paired $t$-tests: $p=0.003$ and $p<0.001$).
The random gate shows no statistically detectable difference from direct lifting ($p=0.48$), indicating that abstention and repeated grasp sampling alone do not explain the improvement of the learned gates. With no-lift abstentions included in the denominator, pooled episode success rates are $59.2\%$, $54.8\%$, $38.2\%$, and $49.5\%$ for the tactile, non-tactile, random, and direct-lift policies, respectively, preserving the same ordering as for executed lifts.
Because random-gate acceptance is independent of the tactile predictor score, its 200 executed lifts directly probe predictor discrimination: success was $86.5\%$ ($77/89$) above the threshold and $16.2\%$ ($18/111$) below, supporting that the predictor distinguishes stable from unstable grasps rather than benefiting from abstention alone.

Qualitatively, contact cues from multiple fingers appeared to help identify partial or off-center grasps on elongated or visually ambiguous objects, such as the sunglasses and light bulb, consistent with the tactile gate's gains over the non-tactile gate in \fig{fig:online_grasping}.
However, success remained $40$--$60\%$ on the three heaviest objects, suggesting that locally plausible pre-lift contacts may not ensure sufficient support under load, given that object mass is not an explicit predictor input.

%% file: 7_conclusion.tex
This paper presents an end-to-end framework for predicting and improving dexterous grasp stability from temporal visuo-tactile sensing.
We collected and now publicly release a large dataset of \numgrasps{} dexterous grasps across \numobjects{} objects using a multi-fingered robotic hand equipped with \digittt{} fingertips, recording vision, proprioception, and tactile streams throughout each grasp.
Across modality and encoding studies, together with controlled analyses of tactile spatial resolution and temporal observations, we show that high-resolution, dynamic touch provides a particularly strong stability signal.
When deployed online as a stability gate, the visuo-tactile model improved success among executed lifts by \bestrelvizsuccessgrasp{} percentage points over the non-tactile gate.
Manipulating objects around us almost always starts with grasping them.
Our work takes a step toward more reliable dexterous grasping and provides evidence that rich tactile feedback can play an important role in achieving it.

Our approach suggests several directions for future work.
Better representations of the interaction between hand configuration and distributed fingertip contact, richer visual geometry from multiple views or depth, and improved cross-modal fusion could further enhance modality complementarity.
By incorporating candidate actions as inputs and continually learning from regrasp trials collected during deployment, the stability predictor could be extended into an action-conditioned model~\citep{calandra2018more} that scores and selects reaching and hand-closure trajectories.
The gate could also serve as a proposal-agnostic pre-lift critic for learned grasp planners and guide closed-loop tactile control during lifting.

%% file: 99_acknowledgments.tex
We thank the ZIH at TU Dresden for providing computing resources.
We acknowledge the use of Claude Code and Codex for coding and debugging (Secs.~\ref{sec:approach}--\ref{sec:result}) and data visualization and figure preparation (Sec.~\ref{sec:result}).
All AI-assisted outputs were reviewed and verified by the authors.

\textbf{Author Contributions.}
R.C. and K.N. conceptualized the study, developed the methodology, supervised and administered the project, and wrote the manuscript.
K.N., A.B., and P.K. collected and curated the data. 
K.N. developed the software, conducted the experiments, formally analyzed the data, validated the results, and prepared the visualizations. 
R.C. provided resources and acquired funding.

%% file: 100_appendix.tex
\section{Large-Scale Dataset Details}

\sec{sec:data} describes how the \numgrasps{} grasp trials were collected and labeled; here we give the details needed to interpret and reuse the dataset.
We report how the \numobjects{} objects are distributed in mass, size, and material, quantify how closely the automatic stability labels match manual verification, and describe what each trial records beyond the grasp-time window used by the predictor in this paper.

\subsection{Object Set}
\begin{figure}[t]
    \centering
    \begin{subfigure}[b]{0.48\textwidth}
        \centering
        \includegraphics[width=\textwidth]{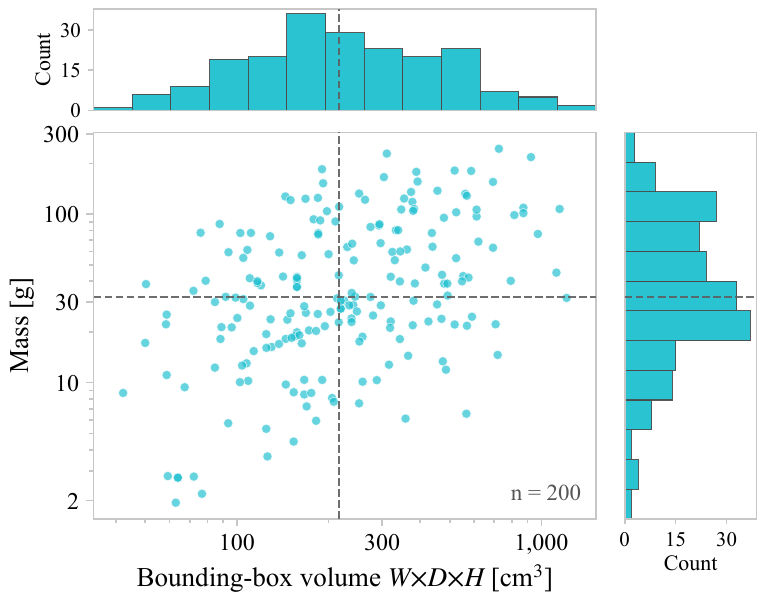}
        \caption{Mass and bounding-box volume}
        \label{fig:appendix_object_set_prop_a}
    \end{subfigure}
    \begin{subfigure}[b]{0.48\textwidth}
        \centering
        \includegraphics[width=\textwidth]{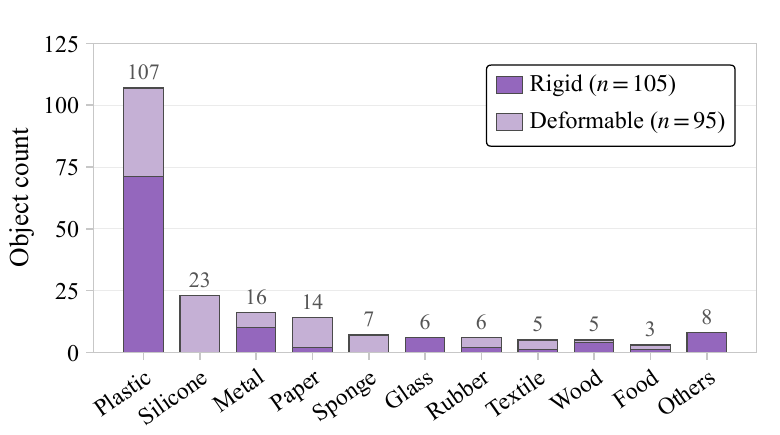}
        \caption{Primary material and compliance}
        \label{fig:appendix_object_set_prop_b}
    \end{subfigure}
    \caption{Physical-property and material distributions of the \numobjects{} dataset objects. (a) Each point represents one object, positioned by its measured mass and dimension-derived bounding-box volume; both axes use logarithmic scales, the top and right histograms show the marginal distributions, and the dashed lines mark their medians. (b) Object counts per primary material, split into rigid and deformable classes; primary materials represented by at most two objects (e.g., leather, ceramic, and stone) are grouped as Others.}
    \label{fig:appendix_object_set_prop}
\end{figure}

The \numobjects{} objects used for grasp data collection (\fig{fig:object_set}) cover variation in size, shape, material, compliance, and visual appearance, exposing the stability predictor to diverse contact conditions rather than repeated grasps of a few canonical objects.
As quantified in \fig{fig:appendix_object_set_prop_a}, object masses span \SIrange[range-phrase={--},range-units=single]{1.94}{244.4}{\gram} (median \SI{32.34}{\gram}), and the dimension-derived bounding-box volumes $W\times D\times H$ span \SIrange[range-phrase={--},range-units=single]{42.30}{1210}{\cubic\centi\meter} (median \SI{216.30}{\cubic\centi\meter}).
Here, we use bounding-box volume as a consistent proxy for object scale rather than physical material volume, which is ambiguous for hollow and irregular objects.
The objects span 16 primary materials and are nearly balanced between rigid (105) and deformable (95) objects, with plastic (107 objects), silicone (23), and metal (16) forming the largest material groups, as detailed in \fig{fig:appendix_object_set_prop_b}.
For the cross-validation in \sec{sec:cross_validation}, object identities are kept disjoint between training and validation folds.
The 20 unseen test objects used for model-guided grasping in \fig{fig:online_grasping} are not among the \numobjects{} objects in \fig{fig:object_set}.

\subsection{Grasp Stability Labeling}
Each grasp trial carries a binary stability label $o \in \{0,1\}$, where $o=1$ denotes a successful lift without dropping the object.
As described in \sec{sec:data}, the robot lifts the object by \SI{100}{\milli\metre} and holds it for \SI{3}{\second} after hand closure, and the system then labels the trial automatically from the pre- to post-lift change in object height.
The height is estimated from SAM~3~\citep{carion2026sam3} segmentation mask, the depth image, and the calibrated camera extrinsics, and the trial is labeled stable when the post-lift height exceeds the pre-grasp height by at least \SI{50}{\milli\metre}.
When SAM~3 fails to detect the object after lifting, e.g., due to hand occlusion, the label falls back to a logistic model over fingertip displacements inferred from the difference between commanded and observed 16-DoF hand configurations.
For the training and evaluation set, we also manually verify the post-lift images and store a corrected binary label for each retained trial.
Treating the automatic labels as predictions and the manually verified labels as ground truth, the pipeline reaches $92.34\%$ agreement ($9234/10000$ trials; \tab{tab:appendix_label_agreement}), a sanity check on the detector and the fallback.

Nevertheless, purely visual post-lift detection can be unreliable under hand occlusion, and the proprioception-based fallback remains tied to the particular object set and robot hand.
In follow-up work on the same recordings, a learned post-lift detector that fuses proprioception, tactile video, contact audio, and external vision reaches $99.14\%$ agreement with the manually verified labels under 5-fold object-disjoint cross-validation.
Restricted to sensing local to the hand (proprioception, tactile video, and contact audio), it still reaches $98.60\%$, and even proprioception alone gives $97.75\%$.
Accurate automatic labeling therefore need not depend on an external camera view or its calibration, and such a detector can be reused when the workspace or surrounding equipment changes.
Future work could extend the labels beyond binary lift success using such local signals, for example by marking slip onset or contact loss during the lift and hold.

\subsection{Multimodal Full-Grasp Recordings}
Although the stability predictor in this paper uses only a grasp-time observation window, each trial continuously records the full grasp attempt, from reaching and hand closure to lifting and post-lift outcome checking.
Each recording spans about \SI{12}{\second}, varying with the reaching motion, and contains external RGB-D observations, robot proprioception, four-fingertip \digittt{} streams, commanded arm and hand actions, phase annotations, and the final stability label.
During hand closing, the 16 joint targets are sampled independently and uniformly from the calibrated ranges in \tab{tab:hand_closing_ranges}, and the controller sends 50 linearly interpolated position-control waypoints from the measured hand pose at a nominal \SI{50}{\hertz}.
Commanded and observed joint states and their tracking errors are recorded at \SI{30}{\hertz}, but neither tracking error nor tactile feedback alters the commanded trajectory: hand closing is open loop at the task level, while contact still changes the observed joint and fingertip-sensor sequences.
Because observations and actions are recorded asynchronously with phase start times, any portion of the grasp can be resampled into a synchronized dataset; the grasp-time window used in this paper is one such instantiation.
Beyond the predictor studied here, these full-grasp recordings can support broader studies of contact-rich grasping, such as action-conditional stability predictors, contact dynamics models, and multimodal representation learning from temporal visual, proprioceptive, and tactile feedback.

\begin{table}[t]
    \centering
    \caption{Confusion matrix of the automatic labels against the manually verified labels, which agree on $92.34\%$ of the \numgrasps{} trials.}
    \vspace{0.5em}
    \label{tab:appendix_label_agreement}
    \footnotesize
    \setlength{\tabcolsep}{6.0pt}
    \renewcommand{\arraystretch}{1.12}
    \begin{tabular}{ll|cc|c|}
        \multicolumn{2}{c|}{} & \multicolumn{2}{c|}{Manual label} & \\
        \multicolumn{2}{c|}{} & 1 (Stable) & 0 (Unstable) & Total \\
        \hline
        \multirow{2}{*}{\shortstack{Automatic label}} & 1 (Stable) & 4361 & 415 & 4776 \\
        & 0 (Unstable) & 351 & 4873 & 5224 \\
        \hline
        & Total & 4712 & 5288 & 10000 \\
        \hline
    \end{tabular}
\end{table}

\begin{table}[t]
    \centering
    \caption{Calibrated per-joint ranges from which the 16 hand joint targets are sampled independently and uniformly during data collection. IP, MCP, CMC, DIP, and PIP denote the interphalangeal, metacarpophalangeal, carpometacarpal, distal interphalangeal, and proximal interphalangeal joints, and ABD denotes the abduction degree of freedom.}
    \label{tab:hand_closing_ranges}
    \vspace{0.5em}
    \footnotesize
    \setlength{\tabcolsep}{3.2pt}
    \renewcommand{\arraystretch}{1.08}
    \begin{tabular}{clcc|clcc}
        \hline
        Finger & Joint & Min. [$^\circ$] & Max. [$^\circ$] & Finger & Joint & Min. [$^\circ$] & Max. [$^\circ$] \\
        \hline
        \multirow{4}{*}{Thumb} & IP  & 25 & 45 & \multirow{4}{*}{Index} & DIP & 25 & 45 \\
        & MCP & 25 & 45 & & PIP & 25 & 45 \\
        & ABD & 50 & 70 & & MCP & 45 & 90 \\
        & CMC & 75 & 90 & & ABD & 0 & 15 \\
        \hline
        \multirow{4}{*}{Middle} & DIP & 25 & 45 & \multirow{4}{*}{Ring} & DIP & 25 & 45 \\
        & PIP & 25 & 45 & & PIP & 25 & 45 \\
        & MCP & 45 & 90 & & MCP & 45 & 90 \\
        & ABD & $-10$ & 5 & & ABD & $-20$ & $-5$ \\
        \hline
    \end{tabular}
\end{table}

\section{Touch Input and Fusion Ablations}

This appendix gives the detailed results behind the three touch-pathway components reported in \sec{sec:cross_validation}: which \digittt{} streams enter the touch branch, how tactile-camera frames are preprocessed, and where the four fingertip representations are merged.
Every ablation in this section starts from the full vision, proprioception, and touch setting and changes only the component under study, using the VideoMAEv2-based model described in \sec{sec:end2end_prediction} and the same 5-fold object-disjoint cross-validation protocol.

\subsection{\digittt{} Submodalities}
We further analyze which \digittt{} streams contribute to grasp stability prediction.
As shown in \tab{tab:digit360_ablation}, the tactile camera nearly matches the full \digittt{} configuration ($83.41\%$ vs. $83.55\%$), while audio, IMU, and pressure stay close to the non-tactile baseline  (Vision + Proprio in \fig{fig:validation_results}), indicating that optical tactile images processed by pretrained video encoders provide the dominant signal in this setting.
Whether these non-camera tactile streams are uninformative here or underused by our shared encoder and fusion design remains open, and encoders matched to each signal may be needed to exploit any complementary cues they carry.
Sparsh-X~\citep{higuera2025tactile} provides such encoders, pretrained across the \digittt{} image, audio, IMU, and pressure streams, but we did not include it here because its pretraining uses modality-specific temporal windows, whereas our predictor consumes a synchronized three-second grasp window across modalities.
It could still serve as a local per-fingertip encoder over sliding windows sampled within the grasp phase, with the resulting embeddings aggregated by our finger-temporal and cross-modal fusion modules.

\subsection{Tactile-Camera Preprocessing}
We compare two tactile-camera preprocessing variants: raw RGB frames and background-subtracted frames.
The background-subtracted variant replaces each tactile frame, per fingertip, with its absolute RGB difference from the first frame of the corresponding grasp-window clip, with the aim of emphasizing contact-induced changes and the contact area.
Raw frames nevertheless perform slightly better ($83.55 \pm 1.58\%$ vs. $82.58 \pm 1.94\%$), consistent with the pretraining of VideoMAEv2 on large-scale RGB videos~\citep{wang2023videomaev2}.
They preserve color, illumination, and contact-induced appearance changes in a format closer to the pretrained backbone, whereas background subtraction emphasizes contact residuals but alters the RGB statistics and can remove stable appearance cues from the gel and illumination field.

\begin{table}[t]
    \centering
    \caption{Mean accuracy with SD over 5-fold CV for \digittt{} submodality ablations in the touch branch, with vision and proprioception retained.}
    \label{tab:digit360_ablation}
    \vspace{0.5em}
    \footnotesize
    \setlength{\tabcolsep}{3.5pt}
    \begin{tabular}{lr}
        \hline
        \digittt{} inputs & Acc. [\%] \\
        \hline
        Full modalities & $83.55 \pm 1.58$ \\
        Camera-only & $83.41 \pm 1.30$ \\
        Audio-only & $79.60 \pm 1.46$ \\
        IMU-only & $79.94 \pm 1.50$ \\
        Pressure-only & $79.93 \pm 1.10$ \\
        No tactile streams (i.e., Vision + Proprio) & $79.56 \pm 1.60$ \\
        \hline
    \end{tabular}
\end{table}

\begin{table*}[t]
    \centering
    \caption{Mean accuracy with SD over 5-fold CV when varying which fingers provide input. Finger-specific inputs (joint angles in every setting, and \digittt{} streams when touch is enabled) are restricted to the listed fingers, while vision is not finger-specific and remains unchanged; all grasps are still executed with the full four-fingered hand.}
    \label{tab:finger_ablation_multimodal}
    \vspace{0.5em}
    \footnotesize
    \setlength{\tabcolsep}{3.0pt}
    \begin{tabular}{lcccc}
        \hline
        & \multicolumn{4}{c}{Acc. [\%]} \\
        \cline{2-5}
        \rule[-2.8ex]{0pt}{5.6ex}\raisebox{-0.5\height}{Finger inputs included}
        & \raisebox{-0.5\height}{Proprio}
        & \raisebox{-0.5\height}{\shortstack{Vision +\\[1.5pt] Proprio}}
        & \raisebox{-0.5\height}{\shortstack{Proprio +\\[1.5pt] Touch}}
        & \raisebox{-0.5\height}{\shortstack{Vision + Proprio\\[1.5pt] + Touch}} \\
        \hline
        Thumb + Index + Middle + Ring
        & $79.99 \pm 0.78$
        & $79.56 \pm 1.60$
        & $82.34 \pm 2.26$
        & $83.55 \pm 1.58$ \\
        Thumb + Index + Middle
        & $79.48 \pm 1.45$
        & $79.34 \pm 0.88$
        & $82.23 \pm 1.80$
        & $83.31 \pm 1.41$ \\
        Thumb + Index
        & $79.00 \pm 0.86$
        & $79.93 \pm 1.50$
        & $81.15 \pm 1.70$
        & $82.63 \pm 1.40$ \\
        Thumb + Middle
        & $79.54 \pm 0.88$
        & $79.83 \pm 1.75$
        & $81.78 \pm 1.52$
        & $83.41 \pm 0.95$ \\
        Thumb
        & $79.38 \pm 1.09$
        & $79.49 \pm 1.24$
        & $81.17 \pm 2.15$
        & $82.55 \pm 1.34$ \\
        Index
        & $68.15 \pm 1.30$
        & $79.59 \pm 1.35$
        & $68.86 \pm 1.09$
        & $79.73 \pm 1.11$ \\
        Middle
        & $72.74 \pm 0.88$
        & $79.72 \pm 1.36$
        & $73.35 \pm 1.01$
        & $81.07 \pm 1.30$ \\
        \hline
    \end{tabular}
\end{table*}

\subsection{Finger Fusion Stage}
We also ablate when the four fingertip representations are merged.
In the global-fusion design described in \sec{sec:end2end_prediction}, the four finger-level touch tokens remain separate until cross-modal fusion, so the fusion tokens can attend to finger-specific contact evidence together with vision and proprioception.
The local-fusion variant instead merges them into a single touch token inside the touch branch.
The two differ by only $0.31$ percentage points ($83.55 \pm 1.58\%$ vs. $83.24 \pm 1.61\%$), suggesting that much of the cross-finger context is already encoded by the factorized Transformer blocks over time and fingers that precede either representation.
A natural extension is to make the fusion more hand-structure-aware, splitting hand proprioception into finger-specific joint-state tokens and fusing each with the corresponding fingertip tactile token before global multimodal aggregation.

\section{Multi-Fingered Embodiment Analyses}

This appendix examines how the multi-fingered embodiment contributes to grasp stability.
The first analysis varies which fingers provide input to the model under the cross-validation protocol of \sec{sec:cross_validation}; the other two are descriptive, relating detected fingertip contacts to grasp success across all \numgrasps{} trials without training a model.

\subsection{Finger-Specific Inputs}
We vary which finger-specific inputs the model receives, restricting joint angles and, when touch is enabled, the corresponding \digittt{} streams to the selected fingers; vision is not finger-specific and stays unchanged, and all grasps are still executed with the full four-fingered hand.
\tab{tab:finger_ablation_multimodal} extends \fig{fig:finger_ablation} with a proprioception-and-touch setting that has no vision.
With proprioception alone, all four fingers reach $79.99\%$, $0.45$ points above the thumb--middle pair, and the thumb alone outperforms the middle finger alone, consistent with the opposable thumb providing the main counter-contact for grasp closure and object support.
Adding vision removes most of this dependence, leaving a spread of $0.59$ points across finger subsets versus $11.84$ without vision, so external observation already recovers much of the hand configuration.
Touch behaves differently: without vision, the all-finger model reaches $82.34\%$ against $81.78\%$ for thumb--middle and $81.15\%$ for thumb--index, and with vision the spread across tactile subsets remains $3.82$ points ($79.73$--$83.55\%$).
Distributed fingertip contact therefore carries stability information that the external camera cannot recover, whereas distributed joint state largely does not.

\subsection{Fingertip Contacts and Grasp Success}
To complement the finger-specific input ablation, we relate the number and identity of detected fingertip contacts at the final pre-lift frame to grasp success.
For each background-subtracted fingertip image, we average the RGB channels and define the contact response as the mean of the brightest $0.1\%$ pixels of the resulting grayscale image.
A contact is detected when this response exceeds a finger-specific threshold manually selected by visual inspection.
We discard detections for which the 3D model of the hand and its fingertip sensors, posed by the measured joint angles at the final pre-lift frame, places the sensor surface within \SI{5}{\milli\metre} of another finger surface, thereby excluding geometric finger--finger self-contact.

\begin{figure*}[t]
    \centering
    \captionsetup[subfigure]{aboveskip=-2pt}
    \begin{subfigure}[b]{0.40\textwidth}
        \centering
        \includegraphics[width=\textwidth]{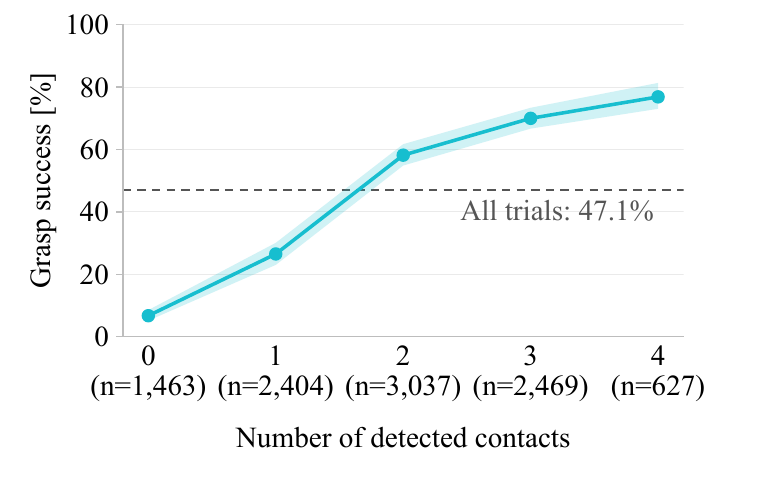}
        \caption{Grasp success by contact count}
        \label{fig:fingertip_contact_analysis_a}
    \end{subfigure}
    \hfill
    \begin{subfigure}[b]{0.59\textwidth}
        \centering
        \includegraphics[width=\textwidth]{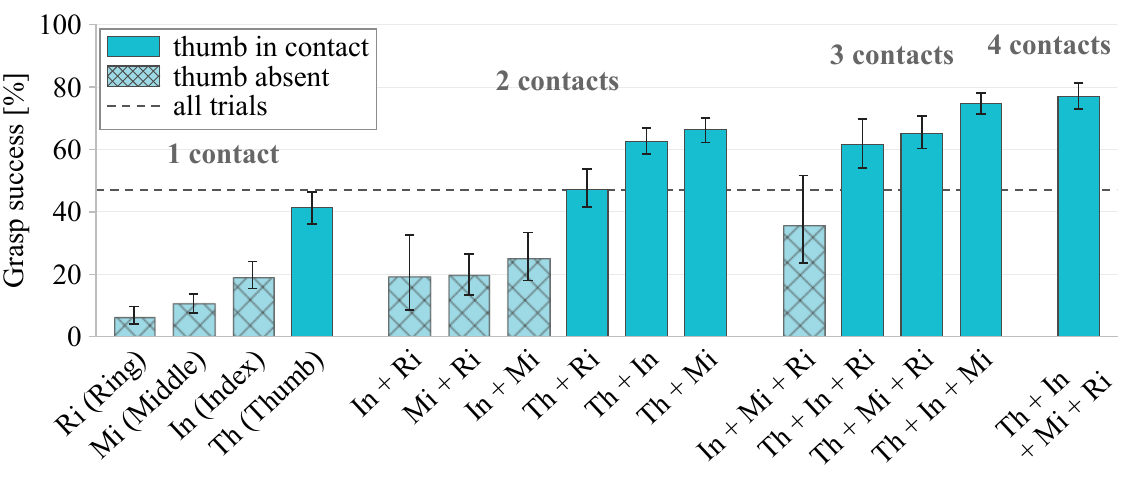}
        \caption{Grasp success by contact pattern}
        \label{fig:fingertip_contact_analysis_b}
    \end{subfigure}
    \caption{Fingertip contacts and grasp success across all \numgrasps{} grasp trials in the dataset. (a) Success rate by detected-contact count; $n$ denotes the number of trials at each contact count. Trials with no or a single detected contact can still succeed when the object is supported outside the fingertip sensing surfaces or when contact falls below the detection threshold. (b) Success rate by non-empty contact set; solid and cross-hatched bars indicate sets with and without the thumb, respectively. Both panels exclude geometric finger--finger self-contact, using a \SI{5}{\milli\metre} clearance threshold on the posed 3D hand model. Shading and error bars show object-cluster bootstrap $95\%$ confidence intervals; the dashed line is the overall success rate.}
    \label{fig:fingertip_contact_analysis}
\end{figure*}

As shown in \fig{fig:fingertip_contact_analysis_a}, success rises monotonically from $6.9\%$ to $26.5\%$, $58.2\%$, $70.0\%$, and $76.8\%$ for zero through four contacts; the largest increase occurs from one to two, and configurations with at least two exceed the overall rate of $47.1\%$.
Because a stable lift generally requires at least two opposing contacts, the successes observed at zero or one detected contact indicate limits of the detector: object contact outside the fingertip sensing surfaces, light contacts that fall below the detection threshold, and inaccuracies in the posed 3D model used for the self-contact exclusion.
Finger identity also matters (\fig{fig:fingertip_contact_analysis_b}): with two contacts, sets containing the thumb reach $47.3$--$66.4\%$ versus $19.6$--$25.2\%$ without it; with three contacts, the same pattern persists at $61.5$--$74.8\%$ versus $35.7\%$.
This is consistent with the opposable thumb providing a principal counter-contact and additional fingertips supplying complementary support.

This analysis is observational: the contacts are detected rather than controlled, so each comparison groups trials that differ in many other ways.
Objects that afford more contacts, or a thumb-opposed contact, may also be easier to lift for reasons unrelated to the contact pattern; the contact labels come from a hand-tuned detector; and rare contact sets rest on few trials, as their wide confidence intervals show.
Still, the pattern links the multi-finger benefit of the input ablation to thumb-opposed contact configurations.

\subsection{Heuristic Contact Baseline}
Analytical force-closure metrics require calibrated contact geometry and force or friction estimates that our setup does not provide.
We therefore compare against a simple embodiment-aware rule built on the same contact detector, following the pattern in \fig{fig:fingertip_contact_analysis_b}: a grasp is predicted stable when the thumb and at least one other fingertip are in contact, excluding geometric finger--finger self-contact.
Against the manually verified labels for all \numgrasps{} trials, this rule reaches $73.07\%$ accuracy ($7307/10000$) with thresholds tuned on the same data.
That is well above the $52.9\%$ majority-class baseline, but below every learned touch-only model in \fig{fig:validation_results}: $78.59\%$ for the static ResNet-50+MLPs encoder, which sees the same single observation as the rule, $79.98\%$ for the ResNet-Transformer, and $82.64\%$ for the VideoMAEv2-based model that encodes the whole grasp window.
A rule based on thumb opposition alone therefore recovers part of the stability signal, and the remaining gap suggests that closing it requires more than binary contact detection: the spatial pattern within each fingertip, how those patterns relate across fingers, and how they evolve during the grasp.